\pdfoutput=1

\documentclass[11pt]{article}

\usepackage{acl}

\usepackage{times}
\usepackage{latexsym}

\usepackage[T1]{fontenc}

\usepackage[utf8]{inputenc}

\usepackage{microtype}
\usepackage{inconsolata}

\usepackage{textcomp}
\usepackage{booktabs} 
\usepackage{multirow}
\usepackage{stfloats}
\usepackage{url}
\usepackage{verbatim}
\usepackage{graphicx}
\usepackage{cite}
\usepackage{bm}
\usepackage{makecell}
\usepackage{color}
\usepackage{arydshln} 
\usepackage{amsmath} 
\usepackage{bbding}
\usepackage{amssymb}
\usepackage{xurl}

\title{Semi-Supervised Spoken Language Glossification}

\author{Huijie Yao$^{1}$~~~~~Wengang Zhou$^{1,}$\thanks{\ \ Corresponding authors: Wengang Zhou and Houqiang~Li}~~~~~Hao Zhou$^{2}$~~~~~Houqiang Li$^{1,}$\footnotemark[1]\\
$^{1}$MoE Key Laboratory of Brain-inspired Intelligent Perception and Cognition, \\
University of Science and Technology of China~~~$^{2}$Baidu Inc. \\
		\texttt{\{yaohuijie, zhouh156\}@mail.ustc.edu.cn, \{zhwg, lihq\}@ustc.edu.cn} \\}

\begin{document}
\maketitle
\begin{abstract}

Spoken language glossification (SLG) aims to translate the spoken language text into the sign language gloss, \emph{i.e.}, a written record of sign language. 
In this work, we present a framework named $S$emi-$S$upervised $S$poken $L$anguage $G$lossification ($S^3$LG) for SLG.
To tackle the bottleneck of limited parallel data in SLG, our $S^3$LG incorporates large-scale monolingual spoken language text into SLG training. 
The proposed framework follows the self-training structure that iteratively annotates and learns from pseudo labels. 
Considering the lexical similarity and syntactic difference between sign language and spoken language, our $S^3$LG adopts both the rule-based heuristic and model-based approach for auto-annotation. 
During training, we randomly mix these complementary synthetic datasets and mark their differences with a special token. 
As the synthetic data may be less quality, the $S^3$LG further leverages consistency regularization to reduce the negative impact of noise in the synthetic data. 
Extensive experiments are conducted on public benchmarks to demonstrate the effectiveness of the $S^3$LG.
Our code is available at \url{https://github.com/yaohj11/S3LG}.
\end{abstract}

\section{Introduction}

Sign Language is the most primary means of communication for the deaf.
Translating between sign and spoken language is an important research topic, which facilitates the communication between the deaf and the hearing~\citep{bragg2019sign,yin-etal-2021-including}.
To support the development of applications, sign language gloss has been widely used as an intermediate step for generating sign language video from spoken language text~\citep{saunders2020progressive,saunders2022signing} or the inverse direction~\citep{pu2020boosting,chen2022two}.
The sign language gloss is the written representation of the signs.
As a generally adopted way for sign language transcription, gloss is sufficient to convey most of the key information in sign language.
In this work, we focus on the first step of the former task named spoken language glossification (SLG), which aims to translate the spoken language text into the sign language gloss.

SLG is typically viewed as a low-resource sequence-to-sequence mapping problem.
The previous methods~\citep{zhu2023neural,walsh2022changing,egea2021syntax,egea2022linguistically} rely on the encoder-decoder architectures~\citep{luong2015effective,sutskever2014sequence} to jointly align the embedding space of both languages in a data-driven manner.
Since the data collection and annotation of sign language requires specialized knowledge, obtaining a large-scale text-gloss dataset is time-consuming and expensive~\citep{de2023machine}.
As a result, the performance of SLG models is limited by the quantity of parallel data~\citep{camgoz2018neural,zhou2021improving}.
Witnessing the success of introducing monolingual data to enhance low-resource translation quality~\citep{cheng2019semi,sennrich2015improving,pan2009survey,zoph2016transfer}  in neural machine translation (NMT), we are motivated to explore the accessible unlabeled spoken language texts to improve SLG.

In this work, we present a framework named $S$emi-$S$upervised $S$poken $L$anguage $G$lossification ($S^3$LG) to boost SLG, which iteratively annotates and learns from pseudo labels.
To implement the above idea, the proposed $S^3$LG adopts both the rule-based heuristic and model-based approach to generate pseudo glosses for unlabeled texts.
The rule-based synthetic data has high semantic accuracy, however, the fixed rules make it difficult to cover complex expression scenarios.
The model-based approach on the other hand is more flexible for learning the correspondence between sign language and spoken language and generates pseudo gloss with higher synthetic diversity.
These complementary synthetic datasets are randomly mixed as a strong supplement for the training of the SLG model.
Besides, the model-based synthetic data is generated by the SLG model, which sets a good stage for iteratively re-training the SLG model.

In addition, $S^3$LG introduces a simple yet efficient design from three aspects.
Firstly, in each iteration, the training process is separated into two stages, \emph{i.e.}, pre-training and fine-tuning for domain adaptation.
Secondly, to encourage the model to learn from the noisy pseudo labels, we apply the consistency regularization term to the training optimization and gradually increase the weight of the consistency regularization in the training curriculum.
It enforces the consistency of the predictions with network perturbations~\citep{gao2021simcse} based on the manifold assumption~\citep{oliver2018realistic}.
Thirdly, to encourage the SLG model to learn complementary knowledge from different types of synthetic data, a special token is added at the beginning of input sentences to inform the SLG model which data is generated by the rule-based or model-based approach~\citep{caswell2019tagged}.
Through end-to-end optimization, our $S^3$LG achieves significant performance improvement over the baseline.
Surprisingly, the experiments show that the translation accuracy on low-frequency glosses is promisingly improved.
We conjecture that the SLG model acts differently in annotating the high-frequency and low-frequency glosses, and such bias is mitigated by the rule-based synthetic data.

In summary, our contributions are three-fold:
\begin{itemize}
\item[$\bullet$] 
We propose a novel framework $S^3$LG for SLG (namely, text-to-gloss translation), which iteratively annotates and learns from the synthetic data.
It adopts two complementary methods, \emph{i.e.}, the rule-based heuristic and model-based approach for auto-annotation.
\end{itemize}
\begin{itemize}
\item[$\bullet$]
We further leverage consistency regularization to reduce the negative impact of noise in the synthetic data.
The biases of the SLG model on low-frequency glosses are mitigated by incorporating the rule-based synthetic data.
\end{itemize}
\begin{itemize}
\item[$\bullet$] 
We conduct extensive experiments to validate our approach, which shows encouraging performance improvement on the two public benchmarks, \emph{i.e.}, CSL-Daily~\citep{zhou2021improving} and PHOENIX14T~\citep{camgoz2018neural}.
\end{itemize}

\section{Related Work} 
In this section, we briefly review the related works on spoken language glossification and semi-supervised learning.

\noindent{\bf Spoken language glossification.}
\citet{camgoz2018neural} publish the first sign language dataset PHOENIX14T and pioneer the linguistic research for sign language~\citep{de2021frozen,cao2022explore}.
With the advance of NMT, the previous methods~\citep{stoll2020text2sign,saunders2020progressive} adopt the encoder-decoder paradigm, which can be specialized using different types of neural networks, \emph{i.e.}, RNNs~\citep{yu2019review}, CNNs~\citep{gehring2017convolutional}.
Considering gloss as a text simplification,~\citet{li2022transcribing} propose a novel editing agent.
Instead of directly generating the sign language gloss, the agent predicts and executes the editing program for the input sentence to obtain the output gloss.
By leveraging the linguistic feature embedding,~\citet{egea2021syntax} achieve remarkable performance improvement.
\citet{egea2022linguistically} further apply the transfer learning strategy result in continues performance increasing.
Recently,~\citet{zhu2023neural} first introduce effective neural machine translation techniques to SLG with outstanding performance improvements, which lays a good foundation for further research.

\noindent{\bf Semi-supervised learning.}
Generating pseudo labels for the unlabeled data is a widely adopted semi-supervised learning algorithm in low-resource NMT, known as back-translation~\citep{gulcehre2015using} and self-training~\citep{zhang2016exploiting}, respectively.
With the target-side monolingual data, back-translation obtains pseudo parallel data by translating the target-side sentences into the source-side sequences.
As an effective data augmentation method, it is widely adopted in the inverse task of SLG, \emph{i.e.}, gloss-to-text translation~\citep{moryossef2021data,zhang2021approaching,angelova2022using,chiruzzo2022translating}.
Due to the lack of the sign language gloss corpus, it is hard to incorporate large-scale monolingual data in the training process of SLG with back-translation~\citep{rasooli-tetrault-2015}.
In contrast, the self-training requires source-side monolingual data to generate pseudo parallel data based on a functional source-to-target translation system.
Since it is hard to optimize a neural translation system with an extremely limited amount of parallel data~\citep{moryossef2021data,zhang2021approaching,xie2020self}, this motivates us to go along this direction and design more effective algorithms.

Different from the aforementioned methods, we focus on iteratively annotating and learning from pseudo labels.
Considering the lexical similarity and syntactic difference between sign language and spoken language, we adopt two complementary approaches (\emph{i.e.}, rule-based heuristic and model-based approach) to generate synthetic data.
Moreover, we put forward the consistency regularization and tagging strategy to reduce the negative impact of noisy synthetic data.

\begin{figure}[t]
\centering
\includegraphics[width=78mm]{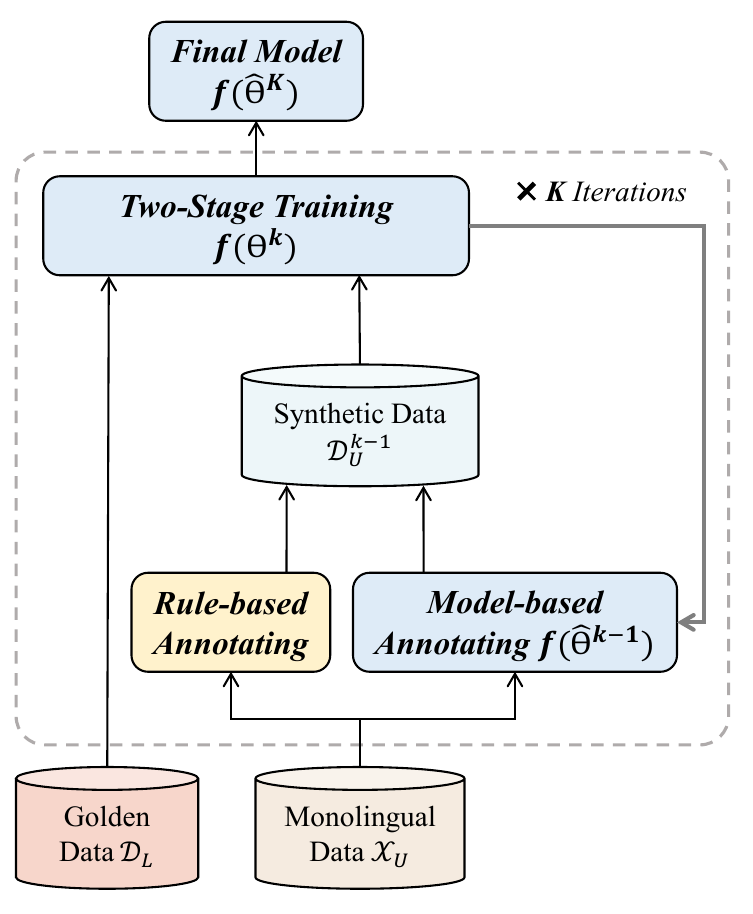}
\caption{
Overview of the proposed $S^3$LG. It iteratively annotates and learns from pseudo labels to obtain the final SLG model $f(\hat{\theta}^K)$ after total $K$ iterations. 
For the $k$-th iteration, the synthetic data $\mathcal{D}_U^{k-1}$ is conducted by randomly mixing the two complementary pseudo glosses generated by the fixed rules and the previously obtained SLG model $f(\hat{\theta}^{k-1})$, respectively.
Note that at the first iteration, only the rule-based synthetic data is available.
}
\label{fig1}
\end{figure}

\section{Methodology}
In this section, we first introduce the overview of our $S^3$LG in Sec.~\ref{overview}; then we elaborate on the annotating methods for monolingual data in Sec.~\ref{Annotating}; and finally, we detail the training strategy in Sec.~\ref{Training}.

\subsection{Overview}~\label{overview}
The primary objective of the SLG model is to acquire knowledge about the mapping $f(\theta): \mathcal{X} \mapsto \mathcal{Y}$, where $\mathcal{X}$ and $\mathcal{Y}$ denote the collection of spoken language text and sign language gloss associated with the vocabulary $\mathcal{V}$, respectively.
$\theta$ is the parameters of the SLG model.
Most SLG model adopts the encoder-decoder architecture, where the input $\bm{x} \in \mathcal{X}$ is first encoded to devise a high-level context representation.
It is then passed to the decoder to generate the output $\bm{y} \in \mathcal{Y}$.
The encoder and decoder can be specialized using different types of neural networks.
Given a set $\mathcal{D}_L=\{(\bm{x}^i_L,\bm{y}^i_L)\}^M_{i=1}$ of $M$ labeled samples and a set $\mathcal{X}_U=\{\bm{x}^i_U\}^N_{i=1}$ of $N$ unlabeled data, we aim to design a semi-supervised framework for SLG to improve text-to-gloss translation by exploring both the labeled and unlabeled data.
To this end, we propose $S^3$LG, which iteratively annotates and learns from two complementary synthetic data generated by the rule-based heuristic and model-based approach, respectively.
 
Fig.~\ref{fig1} provides an overview of the $S^3$LG approach, which consists of three main steps, namely rule-based annotating, model-based annotating, and two-stage training.
At the $k$-th iteration, the synthetic data $\mathcal{D}_U^{k-1}=\{(\bm{x}^i_U,\bm{y}^{i,k-1}_U)\}^N_{i=1}$ is composed of two parts, \emph{i.e.}, rule-based $\mathcal{D}_{U,r}=\{(\bm{x}^i_{U},\bm{y}^{i}_{U,r})\}^N_{i=1}$ and model-based synthetic data $\mathcal{D}_{U,m}^{k-1}=\{(\bm{x}^i_{U},\bm{y}^{i,k-1}_{U,m})\}^N_{i=1}$.
Based on the monolingual data $\mathcal{X}_U=\{\bm{x}^i_U\}^N_{i=1}$, the rule-based and model-based synthetic data are generated by the fixed rules and the functional SLG model $f(\hat{\theta}^{k-1})$ obtained from the previous iterations, respectively.
We randomly mix the two complementary synthetic data and add a special token at the beginning of the input sentences.
The synthetic data $\mathcal{D}_U^{k-1}=\{(\bm{x}^i_U,\bm{y}^{i,k-1}_U)\}^N_{i=1}$ is then concatenated with the original golden data $\mathcal{D}_L=\{(\bm{x}^i_L,\bm{y}^i_L)\}^M_{i=1}$ as a strong supplement for training an SLG model $f(\theta^{k})$, where $N \gg M$.
After repeating $K$ times, we obtain the final SLG $f(\hat{\theta}^{K})$ model.
Notably, in the first iteration, only the rule-based heuristic is available to generate the pseudo gloss sequences for the monolingual data, as in, $\mathcal{D}_U^0=\mathcal{D}_{U,r}$.

\subsection{Annotating Monolingual Data}~\label{Annotating}
Compared with the limited size of text-gloss pairs, the unlabeled spoken language sentences are easy to reach.
To leverage both the labeled and unlabeled data to enhance the SLG performance, we employ the rule-based heuristic and model-based approach to generate the pseudo parallel data and use it to enrich the original golden data for training.

\noindent{\bf Rule-based annotating.}
Given that sign language gloss is annotated based on the lexical elements from the corresponding spoken language, a naive rule is to copy the unlabeled texts as gloss~\citep{zhu2023neural,moryossef2021data}.
Then, we further apply language-specific rules to different sign languages, respectively.
A Chinese spoken language text is first separated at the word level and then one-on-one mapped into the closet glosses based on the lexical similarity.
As German sign language texts often include affixes and markers, thus, we perform lemmatization on each word in the text.
We leverage the open-source spaCy\footnote{\url{https://spacy.io/}}~\citep{honnibal2017spacy} to obtain the linguistics information.
Using the above rule-based annotating system, the monolingual data $\mathcal{X}_U=\{\bm{x}^i_U\}^N_{i=1}$ is mapped as rule-based synthetic data $\mathcal{D}_{U,r}=\{(\bm{x}^i_{U},\bm{y}^{i}_{U,r})\}^N_{i=1}$.
We provide a detailed list of rules in Appendix~\ref{rule}.

\noindent{\bf Model-based annotating.}
While the rule-based heuristic allows high lexical similarity between text and gloss, it cannot capture complicated syntactic divergence between two languages.
Therefore, following the self-training structure, we further employ a functional SLG model to predict the pseudo glosses for monolingual data, based on more flexible correspondence learned from training data.
Because there is a mutually reinforcing relationship between the translation model and the data it generates.
As the model-based synthetic data is generated by the SLG model, it is possible to improve performance by iteratively re-training the SLG model.
At the $k$-th iteration ($k>1$), based on the best SLG model $f(\hat{\theta}^{k-1})$ in $k-1$ iterations, the monolingual data $\mathcal{X}_U=\{\bm{x}^i_U\}^N_{i=1}$ is annotated as model-based synthetic data $\mathcal{D}_{U,m}^{k-1}=\{(\bm{x}^i_U,\bm{y}^{i,k-1}_{U,m})\}^N_{i=1}$.

\subsection{Two-Stage Training}~\label{Training}
\begin{figure}
\centering
\includegraphics[width=75mm]{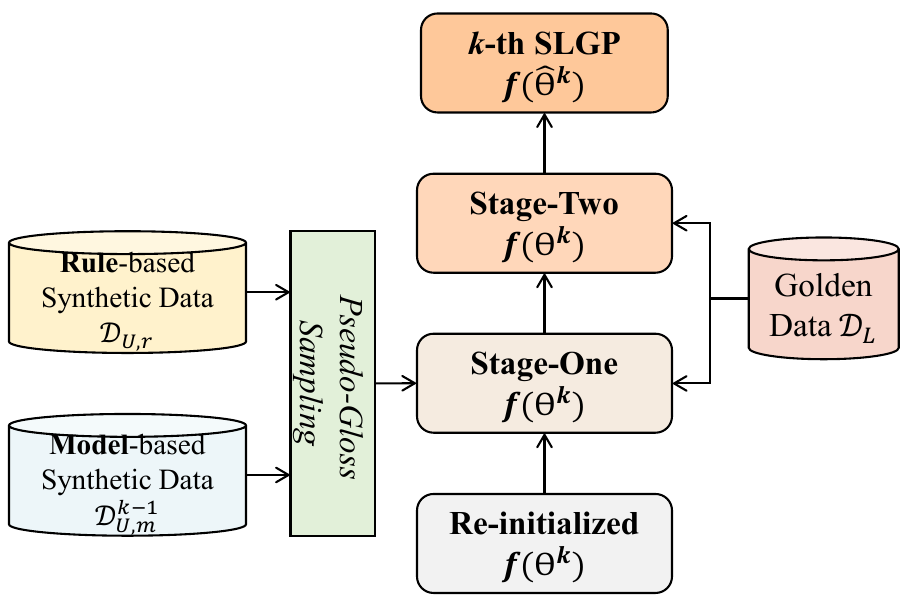}
\caption{
Illustration of the training process in the $k$-th iteration.
}
\label{fig2}
\end{figure}
The proposed $S^3$LG iteratively annotates and learns from the synthetic data.
As $S^3$LG is a data-centric framework, we keep the SLG model simple but competitive, which is the vanilla Transformer model~\citep{vaswani2017attention}.
Without loss of generality, we take the $k$-th iteration as an example to introduce the two-stage training strategy, as shown in Fig.~\ref{fig2}. 

At the beginning of the $k$-th iteration, we re-initialize a new SLG model $f(\theta^k)$, where the input text $\bm{x}=\{x_t\}_{t=1}^{T_x}$ with $T_x$ words is first encoded into a context representation.
The decoder generates the target sequence $\bm{y}=\{y_t\}_{t=1}^{T_y}$ with $T_y$ glosses based on the conditional probability $p(\bm{y}|\bm{x};\theta^k)$.
Specifically, the conditional probability is formulated as:
\begin{equation}
\label{equ:1}
p(\bm{y}|\bm{x};\theta^k)=\prod_{t=1}^{T_{y}}p(\bm{y}_t|\bm{x},\bm{y}_{0:t-1};\theta^k),
\end{equation}
where $\bm{y}_{0:t-1}=\{y_0,\dots,y_{t-1}\}$ denotes the previous output sub-sequence at the $t$-th step.
The initial token $\bm{y}_{0}$ represents the beginning of a sequence.

\subsubsection{Pseudo-Glosses Sampling}
Once the rule-based $\mathcal{D}_{U,r}=\{(\bm{x}^i_U,\bm{y}^{i}_{U,r})\}^N_{i=1}$ and model-based synthetic data $\mathcal{D}_{U,m}^{k-1}=\{(\bm{x}^i_U,\bm{y}^{i,k-1}_{U,m})\}^N_{i=1}$ are obtained, we integrate the annotations at the data level to leverage the two complementary synthetic data.
Aiming at informing the SLG model that the two auto-annotation methods complement each other, a method-specific token is added at the beginning of the spoken language text.
For each unlabeled text $\bm{x}^i_U$, we randomly select a pseudo gloss $\bm{y}^{i,k-1}_U$ from the two pseudo glosses $\bm{y}^{i}_{U,r}$ and $\bm{y}^{i,k-1}_{U,m}$ with equal probability.
Based on the previous $k-1$ iterations, we obtain the synthetic data $\mathcal{D}_{U}^{k-1}=\{(\bm{x}^i_U,\bm{y}^{i,k-1}_{U})\}^N_{i=1}$ for enlarging the training samples to re-train the SLG model.
At the initial iteration, the synthetic data is formulated as $\mathcal{D}_{U}^{0}=\mathcal{D}_{U,r}=\{(\bm{x}^i_U,\bm{y}^{i}_{U,r})\}^N_{i=1}$.

\subsubsection{Training Objective}
For optimizing the SLG model $f(\theta^k)$ with both the synthetic data $\mathcal{D}_U^{k-1}$ and golden data $\mathcal{D}_L$, we introduce two kinds of training objective, \emph{i.e.}, cross-entropy loss, and consistency regularization.

\noindent{\bf Cross-entropy loss.}
As shown in Equ.~\ref{equ:1}, the SLG generates the target translation based on the conditional probability provided by the decoder. 
The cross-entropy loss computed between the annotation and the output of the decoder, which is formulated as:
\begin{equation}
\label{equ:3}
L_{CE}(\bm{x},\bm{y},\theta^k)=-log~p(\bm{y}|\bm{x};\theta^k),
\end{equation}
where the $\bm{y}$ denotes the gloss annotation.

\noindent{\bf Consistency regularization.}
The data distribution should be under the manifold assumption, which reflects the local smoothness of the decision boundary~\citep{belkin2001laplacian}. 
The consistency regularization is computed between two predictions with various perturbations to conform to the manifold assumption.
We apply the network dropout as the perturbation.
With the dropout strategy, the activated parts of the same model are different during training.
The consistency regularization is formulated as:
\begin{equation}~\label{equ:4}
\begin{aligned}
    L_{CR}(\bm{x},\theta^k) =&
    KL(f(\bm{x};\theta_1),f(\bm{x};\theta_2)) \\
    &+KL(f(\bm{x};\theta_2),f(\bm{x};\theta_1)), \\
\end{aligned}
\end{equation}
where $\theta_1$ and $\theta_2$ denotes the different sub-models of the SLG $f(\theta^k)$ with dropout during training.
$f(\bm{x};\theta)$ denotes the predictions given by the SLG model $f(\theta)$.
$KL(teacher, student)$ denotes the KL (Kullback-Leibler) divergence loss that aligns the student’s network to the teacher’s network.

Overall, the loss function of the proposed $S^3$LG is formulated as:
\begin{equation}
\label{equ:5}
L(\bm{x},\bm{y},\theta^k)=L_{CE}(\bm{x},\bm{y},\theta^k)+w\cdot L_{CR}(\bm{x},\theta^k),
\end{equation}
where the weight $w$ balances the effect of two parts of restraints.

\subsubsection{Stage-One and Stage-Two}
To alleviate the domain mismatching between the monolingual data $\mathcal{X}_U$ and golden data $\mathcal{D}_L$, the training process is separated into two stages, \emph{i.e.}, pre-training and fine-tuning, which is a conventional way for domain adaption.
The SLG model is first trained on the concatenation of large-scale synthetic data $\mathcal{D}_U^{k-1}$ and golden data $\mathcal{D}_L$ with the pre-training epochs $T$.
To amplify the impact of synthetic data, the pre-training epochs gradually increase as the iteration grows.
Thus the training objective of stage one is formulated as:
\begin{equation}
\label{equ:6}
\mathop{min}\limits_{\theta^k} \mathop{\sum}\limits_{(\bm{x},\bm{y})\in \mathcal{D}_L\cup\mathcal{D}_U^{k-1}} L(\bm{x},\bm{y},\theta^k).
\end{equation}
Subsequently, this model is fine-tuned only on the low-resource in-domain golden data $\mathcal{D}_L$ until convergence.
The training objective of stage two is formulated as:
\begin{equation}
\label{equ:7}
\mathop{min}\limits_{\theta^k} \mathop{\sum}\limits_{(\bm{x},\bm{y})\in \mathcal{D}_L} L(\bm{x},\bm{y},\theta^k).
\end{equation}

\setlength{\tabcolsep}{3.5pt}
\begin{table*}[!htb]
    \centering
    \resizebox{1.0\linewidth}{!}{
    \resizebox{\textwidth}{!}{
    \begin{tabular}{lccccccccccc}
    \toprule
                                            &\multicolumn{5}{c}{Dev}&\,\,\,& \multicolumn{5}{c}{Test} \\
                                            & ROUGE &BLEU-1 &BLEU-2&BLEU-3&BLEU-4& &  ROUGE &BLEU-1 &BLEU-2&BLEU-3&BLEU-4 \\
         \midrule
        \citet{stoll2020text2sign}          & 48.42 & 50.15 & 32.47 & 22.30 & 16.34 & & 48.10 & 50.67 & 32.25 & 21.54 & 15.26 \\
        \citet{saunders2020progressive}     & 55.41 & 55.65 & 38.21 & 27.36 & 20.23 & & 54.55 & 55.18 & 37.10 & 26.24 & 19.10 \\
        \citet{amin2021sign}                & - & - & - & - & - & & 42.96 & 43.90 & 26.33 & 16.16 & 10.42  \\        
        \citet{egea2021syntax}              & - & - & - & - & - & & - & - & - & - & 13.13 \\ 
        \citet{zhang2021approaching}        & - & - & - & - & - & & - & - & - & - & 16.43 \\
        \citet{li2022transcribing}          & - & - & - & - & - & & 49.91 & - & - & 25.51 & 18.89 \\
        \citet{saunders2022signing}         & 57.25 & - & - & - & 21.93 & & 56.63 & - & - & - & 20.08 \\
        \citet{egea2022linguistically}      & - & - & - & - & - & & - & - & - & - & 20.57 \\
        \citet{walsh2022changing}           & 58.82 & 60.04 & 42.85 & 32.18 & 25.09 & & 56.55 & 58.74 & 40.86 & 30.24 & 23.19 \\
        \citet{zhu2023neural}               & - & - & - & - & 27.62 & & - & - & - & - & 24.89 \\
        \midrule
        Baseline                            & 57.06 & 58.84 & 41.07 & 29.76 & 22.29 & & 55.28 & 57.36 & 38.90 & 27.80 & 20.22 \\
        $S^3$LG                            & \textbf{61.60} & \textbf{62.36} & \textbf{46.30} & \textbf{35.63} & \textbf{28.24} & & \textbf{59.62} & \textbf{60.67} & \textbf{43.69} & \textbf{32.91} & \textbf{25.70} \\
         
         \bottomrule
    \end{tabular}
    }}
    \caption{Performance comparison of our proposed $S^3$LG with methods for SLG on PHOENIX14T.}
    \label{tab:ph}
\end{table*}

\setlength{\tabcolsep}{4.0pt}
\begin{table*}[!htb]
    \centering
    \resizebox{1.0\linewidth}{!}{
    \resizebox{\textwidth}{!}{
    \begin{tabular}{lccccccccccc}
    \toprule
                                            &\multicolumn{5}{c}{Dev}&\,\,\,& \multicolumn{5}{c}{Test} \\
                                            & ROUGE &BLEU-1 &BLEU-2&BLEU-3&BLEU-4& &  ROUGE &BLEU-1 &BLEU-2&BLEU-3&BLEU-4 \\
         \midrule
        \citet{li2022transcribing}            & - & - & - & - & - & & 52.78 & - & - & 29.70 & 21.30  \\
        \midrule
        Baseline                            & 50.26 & 52.90 & 32.79 & 20.94 & 14.05 & & 50.75 & 53.23 & 33.24 & 21.21 & 13.92 \\
        $S^3$LG                            & \textbf{61.52} & \textbf{65.88} & \textbf{47.67} & \textbf{36.05} & \textbf{27.95} & & \textbf{61.75} & \textbf{65.88} & \textbf{47.90} & \textbf{36.06} & \textbf{27.74} \\
         \bottomrule
    \end{tabular}
    }}
    \caption{Performance comparison of our proposed $S^3$LG with methods for SLG on CSL-Daily.}
    \label{tab:csl}
\end{table*}

\section{Experiments}

\subsection{Experimental Setup}
\noindent{\bf Datasets.}
We evaluate our approach on two public sign language translation datasets, \emph{i.e.}, PHOENIX14T~\citep{camgoz2018neural} and CSL-Daily~\citep{zhou2021improving}.
Both datasets provide the original sign language video, sign language gloss, and spoken language text annotated by human sign language translators.
We focus on annotated text-gloss parallel data in this work.
The PHOENIX14T dataset is collected from the German weather forecasting news on a TV station.
The CSL-Daily dataset is a Chinese sign language dataset and covers a wide range of topics in daily conversation.

\noindent{\bf Monolingual data.}
Following the previous work~\citep{zhou2021improving}, we obtain the monolingual spoken language sentences close to the topic of golden data.
For the PHOENIX14T and CSL-Daily dataset, we collect $566,682$ and $212,247$ sentences.
The statistics of the data mentioned above are shown in Appendix~\ref{dataset} and \ref{training data}.

\noindent{\bf Evaluation metrics.}
Referring to the previous works~\citep{zhou2021improving,li2022transcribing,zhu2023neural}, we quantify the performance of the generated gloss in terms of accuracy and consistency based on the BLEU~\citep{papineni2002bleu} and ROUGE~\citep{rouge2004package}, respectively.
The BLEU-$N$ ($N$ ranges from $1$ to $4$) is widely used in NMT to show the matching degree of $N$ units between two sequences.
Besides, the ROUGE cares more about the fluency degree of generated sequences.
Both the evaluation metrics indicate attributes of generated gloss, noting that the higher values demonstrate better translation performance.

\noindent{\bf Training settings.}
We implement the proposed approach on Pytorch~\citep{paszke2017automatic}.
For PHOENIX14T and CSL-Daily, the SLG model consists of $5$ and $3$ layers, respectively.
To mitigate the overfitting problem, we apply the common strategy such as dropout and label smoothing.
For the network setting, the dimensions of the embedding layer and the feed-forward network are $512$ and $2048$, respectively.
The number of the attention head is $8$.
For the optimization setting, we leverage the Adam~\citep{kingma2014adam}.
During training, the learning rate and batch size are fixed to $5\times10^{-5}$ and $32$, respectively.
As at the beginning, the predictions of the SLG model might be unreliable, the consistency regularization weight $w$ ramps up, starting from zero, along a linear curve until reaching the $w$~\citep{laine2016temporal}.
Following the previous setting~\citep{li2019understanding}, we randomly shuffle and drop words of the spoken language sentences as data augmentation.

\noindent{\bf Inference details.}
In inference, we use the beam search strategy~\citep{wu2016bridging} to increase the decoding accuracy.
For both the CSL-Daily and the PHOENIX14T dataset, the search width and length penalty are set to $3$ and $1.0$, respectively.
In the process of generating the pseudo glosses for monolingual data, we simply set the search width to $1$ for efficiency.
The experiment is run on an NVIDIA GeForce RTX 3090 with approximately $40$ hours of computational time.

\subsection{Comparison with State-of-the-Art Methods}
We compare the proposed $S^3$LG with the previous text-to-gloss systems on two public benchmarks, \emph{i.e.}, PHOENIX14T~\citep{camgoz2018neural} and CSL-Daily~\citep{zhou2021improving}.
The performances are shown in Tab.~\ref{tab:ph} and Tab.~\ref{tab:csl}, respectively.

As our goal is to explore how to incorporate monolingual data for SLG, our baseline adopts the vallia Transformer as the SLG model which only learns from the original golden data.
By combining all proposed components, our $S^3$LG achieves substantial improvements against the baseline across all evaluation metrics.
The $S^3$LG achieves $28.24$ and $27.95$ BLEU-4 on the DEV set of the PHOENIX14T and CSL-Daily dataset, which surpasses the baseline by $5.95$ and $13.9$, respectively.
The quantitative results demonstrate the effectiveness of utilizing the complementary synthetic data and designs in our $S^3$LG.
For the PHOENIX14T and CSL-Daily datasets, we evaluate the performance using the gloss-level tokenizer as the original annotations, respectively.

\citet{zhu2023neural} provides translation performance in different settings, including semi-supervised, transfer learning, and multilingual.
For a fair comparison, we cite their best performance in the monolingual and bilingual settings.
The results prove the advantage of our novel designs, which distinguishes our approach from previous SLG systems.
The previous works mainly tested on the PHOENIX14T datasets, while CSL-Daily is also an important benchmark for different sign language tasks.
To attract more research attention on Chinese Sign Language, we report our performance on this dataset.

\setlength{\tabcolsep}{1pt}
    \begin{table}[!t]
     \scriptsize
     \resizebox{1.0\linewidth}{!}{
     \begin{tabular}{lccccc}
     \toprule
         Setting         & ROUGE &BLEU-1 &BLEU-2&BLEU-3&BLEU-4\\
         \midrule
         Baseline                  & 57.06 & 58.84 & 41.07 & 29.76 & 22.29 \\
         +~Self-training           & 58.31 & 60.90 & 42.39 & 30.75 & 22.97 \\
         +~Consistency             & 59.74 & 60.14 & 43.54 & 32.64 & 25.25 \\
         +~Rule-based              & 59.86 & 60.01 & 44.65 & 34.25 & 27.12 \\
         +~Aug.+Tag.               & \textbf{61.60} & \textbf{62.39} & \textbf{46.30} & \textbf{35.63} & \textbf{28.24} \\
         \bottomrule
     \end{tabular}
     }
     \caption{Effect of our proposed components.
     `Self-training' represents directly applying the iterative self-training strategy to the baseline.
     `Consistency' denotes adding the additional restraint consistency regularization for the training objective.
     `Rule-based' and `Aug.+Tag.' denote combining the model-based synthetic data with the rule-based synthetic data and marking them with the tagging process, respectively. 
     }
     \label{tab:3}
 \end{table}

\setlength{\tabcolsep}{4.5pt}
    \begin{table}[!t]
     \scriptsize
     \resizebox{1.0\linewidth}{!}{
     \begin{tabular}{lccccc}
     \toprule
         $w$         & ROUGE &BLEU-1 &BLEU-2&BLEU-3&BLEU-4\\
         \midrule
         % 0               &  &  &  &  &  \\
         1               & 60.22 & 62.99 & 45.40 & 34.34 & 26.82 \\
         5               & 61.58 & \textbf{64.10} & \textbf{47.08} & 35.60 & 27.75 \\
         20              & \textbf{61.60} & 62.36 & 46.30 & \textbf{35.63} & \textbf{28.24} \\
         40              & 59.04 & 57.31 & 42.43 & 32.31 & 25.36 \\
         \bottomrule
     \end{tabular}
     }
     \caption{Impact of the consistency regularization weight $w$.
     }
     \label{tab:4}
 \end{table}
   
\subsection{Ablation Study}
To validate the effectiveness of each component proposed in our $S^3$LG framework, unless otherwise specified, we put forward several ablation studies on the DEV set of the PHOENIX14T dataset.

\noindent{\bf Impact of proposed components.}
The main difference between our proposed method and the existing works is to leverage the complementary synthetic as a supplement for the training of the SLG model.
To evaluate the effectiveness of each proposed component, we gradually add them to the baseline SLG system.
Directly applying the iterative self-training process to the baseline delivers a performance gain of $0.68$ BLEU-4, which motivates us to design a more effective algorithm.
We further apply consistency regularization to enforce the predictions of synthetic data under the manifold assumption, which achieves a gain of $2.28$ BLEU-4.
Subsequently, combining the rule-based and model-based synthetic data can be helpful with $1.87$ BLEU-4 improvements.
Besides, the result shows that tagging and applying data augmentation to different types of synthetic data is also a useful strategy, which provides a further gain of $1.12$ BLEU-4.
The results are shown in Tab.~\ref{tab:3}.

\noindent{\bf Impact of $w$.}
In our experiments, the consistency regularization weight $w$ is set to $20$.
This hyper-parameter determines the importance of the consistency regularization compared with the cross-entropy loss.
In Tab.~\ref{tab:4}, we examine the effect of consistency regularization weight with a set of different values.
As the $w$ is set to be $20$, $S^3$LG achieves its best performance.

\setlength{\tabcolsep}{3.5pt}
    \begin{table}[!t]
     \scriptsize
     \resizebox{1.0\linewidth}{!}{
     \begin{tabular}{lccccc}
     \toprule
         $K$         & ROUGE &BLEU-1 &BLEU-2&BLEU-3&BLEU-4\\
         \midrule
         0               & 57.06 & 58.84 & 41.07 & 29.76 & 22.29 \\
         1               & 60.66 & 60.94 & 45.02 & 34.37 & 27.12 \\
         2               & 61.16 & 62.21 & 45.79 & 34.95 & 27.80 \\
         3               & 61.24 & 62.30 & 46.19 & 35.42 & 28.06 \\
         4               & \textbf{61.60} & \textbf{62.36} & \textbf{46.30} & \textbf{35.63} & \textbf{28.24} \\
         5               & 61.60 & 62.36 & 46.30 & 35.63 & 28.24 \\
         \bottomrule
     \end{tabular}
     }
     \caption{Impact of iteration number $K$.
     `0' and `1' denote the baseline and only applying the rule-based synthetic data to enlarge the training data, respectively.
     }
     \label{tab:5}
 \end{table}

\setlength{\tabcolsep}{3pt}
    \begin{table}[!t]
     \scriptsize
     \resizebox{1.0\linewidth}{!}{
     \begin{tabular}{rccccc}
     \toprule
         Scale          & ROUGE &BLEU-1 &BLEU-2&BLEU-3&BLEU-4\\
         \midrule
         0$\times$                 & 57.06 & 58.84 & 41.07 & 29.76 & 22.29 \\
         5$\times$                 & 60.96 & 61.00 & 44.90 & 34.18 & 26.91 \\ 
         % 10$\times$                & 60.98 & 61.69 & 45.40 & 34.50 & 26.96 \\
         15$\times$                & 60.94 & 61.63 & 45.88 & 35.27 & 27.98 \\
         >30$\times$  & \textbf{61.60} & \textbf{62.36} & \textbf{46.30} & \textbf{35.63} & \textbf{28.24} \\
         \bottomrule
     \end{tabular}
     }
     \caption{Scale of synthetic data.
    `0$\times$' and `>30$\times$' represent utilizing non-monolingual data and all the monolingual data for the training of the SLG model, respectively.
     }
     \label{tab:6}
 \end{table}

\noindent{\bf Impact of iteration number $K$.}
The iteration number $K$ is an important hyper-parameter and fixed to $4$ in the previous experiments.
To explore the effect of $K$, we conduct experiments with different iteration numbers.
Tab.~\ref{tab:5} shows the best performance of the $K$ iterations, where $K=1$ (\emph{i.e.}, the initial iteration), the training data is composed of the rule-based synthetic data and golden data.
By combining both the rule-based and model-based synthetic data, the performance of the SLG model is converged at the $4$-th iteration.

\setlength{\tabcolsep}{2.5pt}
    \begin{table}[!t]
     \scriptsize
     \resizebox{1.0\linewidth}{!}{
     \begin{tabular}{llcccc}
     \toprule
         Quantity               & Method & ROUGE &BLEU-2&BLEU-3&BLEU-4\\
         \midrule
         \multirow{2}{*}{1\%}      & Baseline    & 17.63 & 8.31 & 4.86 & 3.20 \\
                                   & Ours        & \textbf{33.44} & \textbf{17.29} & \textbf{9.05} & \textbf{4.69} \\
        \midrule
        \multirow{2}{*}{5\%}       & Baseline    & 34.20 & 19.44 & 11.16 & 7.24 \\
                                   & Ours        & \textbf{45.14} & \textbf{26.93} & \textbf{17.56} & \textbf{12.03} \\
        \midrule
        \multirow{2}{*}{25\%}      & Baseline    & 49.44 & 33.89 & 22.87 & 16.47 \\
                                   & Ours        & \textbf{55.44} & \textbf{40.01} & \textbf{29.44} & \textbf{22.33} \\
         \bottomrule
     \end{tabular}
     }
     \caption{Quantity of annotated data.
     }
     \label{tab:7}
 \end{table}

\setlength{\tabcolsep}{1.5pt}
    \begin{table}[!t]
     \scriptsize
     \resizebox{1.0\linewidth}{!}{
     \begin{tabular}{lcccccc}
     \toprule
         $T$        &  Growing   & ROUGE &BLEU-1 &BLEU-2&BLEU-3&BLEU-4\\
         \midrule
         0          & \XSolidBrush    & 57.06 & 58.84 & 41.07 & 29.76 & 22.29 \\
         1          & \XSolidBrush    & 59.20 & 58.73 & 42.96 & 32.44 & 25.32 \\
         5          & \XSolidBrush    & 60.52 & 59.88 & 44.22 & 33.80 & 26.72 \\
         10         & \XSolidBrush    & 60.94 & \textbf{61.18} & 45.08 & 34.44 & 27.17 \\
         15         & \XSolidBrush    & \textbf{60.97} & 61.05 & \textbf{45.32} & \textbf{34.77} & 27.37 \\
         30         & \XSolidBrush    & 60.56 & 60.59 & 44.98 & 34.63 & \textbf{27.69} \\
         \midrule
         15         & \checkmark   & 61.60 & 62.36 & 46.30 & 35.63 & 28.24 \\
         \bottomrule
     \end{tabular}
     }
     \caption{Effect of pre-training epochs $T$.
     `Growing' represents increasing the pre-training epochs between different iterations.
     }
     \label{tab:8}
 \end{table}

\setlength{\tabcolsep}{1.5pt}
    \begin{table}[!t]
     \scriptsize
     \resizebox{1.0\linewidth}{!}{
     \begin{tabular}{lccccc}
     \toprule
         Setting                & ROUGE & BLEU-1 &BLEU-2&BLEU-3&BLEU-4\\
         \midrule
         Original-Baseline      & 57.06 & 58.84 & 41.07 & 29.76 & 22.29 \\
         ~+$S^3$LP              & 61.60 & 62.36 & 46.30 & 35.63 & 28.24 \\
        \midrule
        BERT-Baseline           & 58.26 & 59.84 & 42.37 & 31.21 & 23.62 \\
        ~+$S^3$LP               & \textbf{62.00} & \textbf{63.72} & \textbf{47.27} & \textbf{36.14} & \textbf{28.48} \\
         \bottomrule
     \end{tabular}
     }
     \caption{Impact of leveraging synthetic data.
     `BERT-Baseline' denotes a stronger baseline by enhancing the original baseline with the pre-trained language model, BERT.
     }
     \label{tab:10}
 \end{table}

\noindent{\bf Scale of synthetic samples.}
As mentioned above, the collected monolingual data outnumbers the annotated data over $30$ times.
In the previous experiments, all the monolingual data is incorporated with the golden data to enhance the training process of the SLG model.
In this ablation study, we investigate the scale of synthetic data.
As shown in Tab.~\ref{tab:6}, the performance improves approximately along a log function curve according to the synthetic data volume.
We speculate that pre-training on too much noisy out-of-domain synthetic data may drown the impact of golden data, which finally causes limited performance improvements.

\noindent{\bf Quantity of annotated golden data.}
As the purpose of our proposed approach is to learn from the monolingual with pseudo glosses, we provide experiments regarding various quantities of synthetic data, in Tab.~\ref{tab:7}.
Under all settings, $S^3$LG outperforms the baseline.
However, the experiment suggests that our proposed approach is more suitable for the scene when the baseline SLG performance is in the range of $7-20$ BLEU-4.

\noindent{\bf Effect of pre-training epochs $T$.}
In Tab.~\ref{tab:8}, we evaluate the impact of pre-training with synthetic data with different epochs.
To simplify the hyperparameter search process, we first conduct the experiments with fixed pre-training epochs $T$ between different iterations and further apply the pre-training epochs growing strategy to it.
The $S^3$LG achieves the best performance under the setting of that 
the pre-training epoch is $15$ for the first iteration and then gradually increases by $10$ between iterations.

\noindent{\bf Impact of leveraging synthetic data.}
To verify whether our performance improvement mainly comes from the synthetic data rather than simply enhancing the encoder with the monolingual data, we conduct experiments by enhancing the original baseline with the pre-trained language model, BERT~\citep{devlin-etal-2019-bert}.
As the results presented in Tab.~\ref{tab:10}, we observe that leveraging the pre-trained language model improves the translation quality, while our proposed approach achieves larger performance gains. We further combine our approach with the pre-trained language model.
The experimental results demonstrate that the performance improvement of our approach stems from two aspects: better comprehension of the encoder and better generation capability of the decoder.

\noindent{\bf Results of CHRF metric.}
To provide more information, following the previous work~\citep{muller-etal-2023-considerations}, we evaluate the performance of our proposed approach based on CHRF~\citep{popovic2016chrf} metric.
The $S^3$LP achieves $56.02$ and $54.84$ CHRF on the DEV and TEST set of PHOENIX14T, which surpasses the baseline ($52.00$ and $51.32$) by $4.02$ and $3.52$, respectively.

\setlength{\tabcolsep}{3.5pt}
    \begin{table}[!t]
     \scriptsize
     \resizebox{1.0\linewidth}{!}{
     \begin{tabular}{lccccc}
     \toprule
        Appearance      & $\leq 3$ & $\leq 6$ &$\leq 8$ & $\leq 10$ & $\leq 15$ \\
         \midrule
         Amount                                 & 42 & 49 & 56 & 57 & 66 \\
         Baseline                               & 2.38 & 1.75 & 1.44 & 2.24 & 8.46 \\
         Model-Based                            & 2.38 & 1.75 & 1.44 & 3.37 & 10.00 \\
         Model \& Rule-based                    & 2.38 & \textbf{3.50} & \textbf{4.34} & \textbf{6.74} & \textbf{15.38} \\
         \bottomrule
     \end{tabular}
     }
     \caption{Translation accuracy of low-frequency glosses.
    `Appearance' represents how many times the glosses appear in the TRAIN set of golden parallel data.
    `Amount' denotes how many samples contain the low-frequency glosses in the DEV set.
     }
     \label{tab:9}
 \end{table}

\noindent{\bf Translation accuracy of low-frequency glosses.}
As the SLG model tends to predict the glosses with high frequency in training data, we believe that utilizing the model-free annotating approach can mitigate the model-based annotating bias.
To verify this, we put forward the experiments for different synthetic data settings with the translation accuracy metric under different low-frequency standards, namely, a gloss appears less than how many times are considered as low frequency, as shown in Tab.~\ref{tab:9}.
The translation accuracy of low-frequency glosses is formulated as $accuracy=N_{pred}/N_{all}$, where $N_{pred}$ and $N_{all}$ denote the number of samples that are predicted with the correct low-frequency glosses and samples that contain the low-frequency glosses, respectively.
We can see that the translation accuracy of the SLG model leveraging both the model-based and rule-based synthetic data achieves promising improvements against the model-based one.

\section{Conclusion}
In this work, we present a semi-supervised framework named $S^3$LG for translating the spoken language text to sign language gloss.
With the goal of incorporating large-scale monolingual spoken language texts into SLG training, we propose the $S^3$LG approach to iteratively annotate and learn from pseudo glosses.
Through a total of $K$ iterations, the final SLG model achieves significant performance improvement against the baseline.
During each iteration, the two complementary synthetic data generated from the rule-based and model-based approaches are randomly mixed and marked with a special token.
We introduce a consistency regularization to enforce the consistency of the predictions with network perturbations.
Extensive ablation studies demonstrate the effectiveness of the above designs.
Besides, the translation accuracy on low-frequency glosses is improved.

\section*{Limitations}
In the hope of attracting more research attention for SLG in the future, we provide the detailed limitations in the next.
On the one hand, the results (see Tab.~\ref{tab:6} and Tab.~\ref{tab:7}) of the experiments suggest that in the extremely low-resource scenarios, our performance improvements might be less significant as a large amount of monolingual data is available.
We conclude that with the very limited golden data as anchors, it is hard to learn from the large-scale synthetic data. 
Although utilizing synthetic data as a strong supplement for training data can achieve promising performance improvements, it is hard to achieve the equal impact of enlarging the training data with annotated data from human interpreters.
To promote the development of sign language research, the fundamental way might be continuously collecting large amounts of annotated data.

On the other hand, we realize that sign language glosses do not properly represent sign languages~\citep{muller-etal-2023-considerations}.
However, as explained in the introduction section, sign language glosses are sufficient to convey most of the key information in sign language.
Given the resource limitations and the current technological capabilities, we believe the two-stage way (\emph{i.e.}, text-to-gloss and gloss-to-gesture) is more achievable and practical to support the development of the application for converting spoken language into sign language.
There are many solutions for animating a 3D avatar and making gloss-indexed gestures smoothly and naturally.
Correspondingly, research on the former stage lacks enough attention.
We think improving the SLG model's performance can be a promising way to implement better sign language production systems.

\section*{Ethical Considerations}
Even with extensive advances in the development of neural machine translation methods in the spoken language area, the study of sign language is still in its infancy.
At the same time, the development of different sign languages is very uneven.
According to the existing approaches, the current researches mainly focus on DGS, leaving other sign languages unexplored.
As studied in~\citep{bragg2019sign,yin-etal-2021-including}, there is limited research to reveal the sign language linguistics character, which also limits the utility of prior knowledge.

As most sign language researchers are hearing people, the provided sign language system might not meet the actual needs of the deaf.
Therefore, to bridge the commutation between the two communities, cooperation could be mutual.
By consulting with native signers, we will proactively seek to design the translation system to be inclusive and user-centered in the future.
We also encourage people of both communities to try out the existing systems and point out their disadvantages to guide the promising direction and accelerate the study of sign language.

\section*{Acknowledgements}
This work is supported by National Natural Science Foundation of China under Contract U20A20183 \& 62021001, and the Youth Innovation Promotion Association CAS. It was also supported by GPU cluster built by MCC Lab of Information Science and Technology Institution, USTC, and the Supercomputing Center of the USTC. 

\bibliographystyle{acl_natbib}
\bibliography{custom}

\appendix
\section{Rules Used in the Rule-based Heuristic for Creating Synthetic Data $\mathcal{D}_{U,r}$}~\label{rule}

\subsection{Chinese Rules}
For a spoken text $\bm{y}$,

1. Build the vocabulary for the gloss $\mathcal{V}$ and spoken word $\mathcal{S}$ from the original golden data $\mathcal{D}_L$. 

2. Tokenize the spoken text at the word-level as $\bm{y}=\{y_1,y_2,\dots,y_T\}$ with $T$ words.

3. Replace the words $y_t \in \bm{y}$ not in the word vocabulary $\mathcal{S}$ by a special token <UNK>.

4. Replace each spoken word $y_t \in \bm{y} $ by the most similar gloss in the gloss vocabulary $\mathcal{V}=\{v_1,v_2,\dots,v_{\left| \mathcal{V} \right|}\}$ based on the lexical similarity, which is formulated as:
\begin{equation}
\label{equ:8}
Sim(y_t,v_i)=E(y_t)\cdot E(v_i)
\end{equation}
where $E(\cdot)$ denotes $L_2$ normalized word embedding processes.
The above information is obtained from the Chinese model (zh\_core\_web\_lg) of spaCy.
 
\subsection{German Rules}
For a spoken text $\bm{y}=\{y_1,y_2,\dots,y_T\}$ with $T$ words,

1. Build the vocabulary for the gloss $\mathcal{V}$ and spoken word $\mathcal{S}$ from the original golden data $\mathcal{D}_L$. 

2. Replace the words $y_t \in \bm{y}$ not in the word vocabulary $\mathcal{S}$ by a special token <UNK>.

3. Lemmatize all the spoken words.

4. Replace the token $y_t \in \bm{y}$ that only matches parts of compounds glosses $v_i \in \mathcal{V}$ by it $v_i$.

The above information is obtained from the German model (de\_core\_news\_lg) of spaCy.

\section{Statistics of Sign Language Datasets}~\label{dataset}
As shown in Tab.~\ref{tab:12} and Tab.~\ref{tab:13}, we present the key statistics of the PHOENIX14T and CSL-Daily dataset, respectively.

The PHOENIX14T dataset is about weather forecasting and does not contain any information that names or uniquely identifies individual people or offensive content.
The CSL-Daily dataset is screened by its publishing team and is about daily life (shopping, school, travel, etc.). Does not contain any information that names or uniquely identifies individual people or offensive content.

 \setlength{\tabcolsep}{2pt}
    \begin{table}[!t]
     \scriptsize
     \resizebox{1.0\linewidth}{!}{
     \begin{tabular}{lccccccc}
     \toprule
                       & \multicolumn{3}{c}{Text} & & \multicolumn{3}{c}{Gloss}  \\
                       & TRAIN & DEV & TEST & & TRAIN & DEV & TEST  \\
         \midrule
         Sentence      & 7,096 & 519 & 642 & & 7,096 & 519 & 642 \\
         Vocabulary    & 2,887 & 951 & 1,001 & & 1,085 & 393 & 411 \\
         Tot. Words    & 99.081 & 6,820 & 7,816 & & 55,247 & 3,748 & 4,264 \\
         Tot. OOVs     & - & 57 & 60 & & - & 14 & 19 \\
         \bottomrule
     \end{tabular}
     }
     \caption{Statistic of the PHOENIX14T dataset.
     }
     \label{tab:12}
 \end{table}
 
\setlength{\tabcolsep}{1pt}
    \begin{table}[!t]
     \scriptsize
     \resizebox{1.0\linewidth}{!}{
     \begin{tabular}{lccccccc}
     \toprule
                            & \multicolumn{3}{c}{Text} & & \multicolumn{3}{c}{Gloss}  \\
                            & TRAIN & DEV & TEST & & TRAIN & DEV & TEST  \\
         \midrule
         Sentence           & 18,401 & 1,077 & 1,176 & & 18,401 & 1,077 & 1,176   \\
         Vocabulary          & 2,343 & 1,358 & 1,358 & & 2,000 & 1,344 & 1,345 \\
         Tot. Words/Chars    & 291,048 & 17,304 & 19,288 & & 133,714 & 8,173 & 9,002 \\
         Tot. OOVs           & - & 64 & 69 & & - & 0 & 0 \\
         \bottomrule
     \end{tabular}
     }
     \caption{Statistic of the CSL-Daily dataset.
     }
     \label{tab:13}
 \end{table}

\section{Statistics of the Training Data}~\label{training data}
The statistics of training data is shown in shown in Tab.~\ref{tab:14}.
To collect more spoken language texts, we extract a subset of CLUE corpus~\citep{xu-etal-2020-clue} based on the topic of daily lives.

In the process of selecting this part of the data, the words in the PHOENIX14T and CSL-Daily datasets are used to select content on related topics, so most of the data is related to weather and daily life, which does not contain any information that names or uniquely identifies individual people or offensive content.

\setlength{\tabcolsep}{1pt}
    \begin{table}[!t]
     \scriptsize
     \resizebox{1.0\linewidth}{!}{
     \begin{tabular}{lcc}
     \toprule
         Type              & Amount & Source \\
         \midrule
         Golden Data                    & 7.096 & PHOENIX14T \\
         Monolingual Data               & 212,247 & Wiki, German weather corpus\\
         \midrule
         Golden Data                    & 18,402 & CSL-Daily \\
         Monolingual Data               & 566,682 & Wiki, CLUE, Web \\
         \bottomrule
     \end{tabular}
     }
     \caption{Statistic of the Training Data.
     }
     \label{tab:14}
 \end{table}

\end{document}